\documentclass[fleqn,10pt]{wlscirep}
\usepackage[utf8]{inputenc}
\usepackage[T1]{fontenc}
\usepackage{multirow}
\usepackage{graphicx}
\usepackage[normalem]{ulem}

\usepackage{algorithm}
\usepackage{algorithmic}
\usepackage{color, soul}
\usepackage{subfigure}
\usepackage{amsmath,amsthm}
\usepackage{bbm}
\usepackage{amssymb}
\usepackage{mathtools}
\usepackage{hhline}
\usepackage{multirow}
\usepackage{newfloat}
\usepackage{listings}
\usepackage{caption}

\theoremstyle{definition}
\newtheorem{definition}{Definition}

\useunder{\uline}{\ul}{}
\title{Defining and Evaluating Decision and Composite Risk in Language Models Applied to Natural Language Inference}
\author[1]{Ke Shen}
\author[1,*]{Mayank Kejriwal}

\affil[1]{University of Southern California, Information Sciences Institute, Marina del Rey, 90292, United States of America}

\affil[*]{kejriwal@isi.edu}

\begin{abstract}
Despite their impressive performance, large language models (LLMs) such as ChatGPT are known to pose important risks. One such set of risks arises from misplaced confidence, whether over-confidence or under-confidence, that the models have in their inference. While the former is well studied, the latter is not, leading to an asymmetry in understanding the comprehensive risk of the model based on misplaced confidence. In this paper, we address this asymmetry by defining two types of risk (decision and composite risk), and proposing an experimental framework consisting of a two-level inference architecture and appropriate metrics for measuring such risks in both discriminative and generative LLMs. The first level relies on a decision rule that determines whether the underlying language model should abstain from inference. The second level (which applies if the model does not abstain) is the model's inference. Detailed experiments on four natural language commonsense reasoning datasets using both an open-source ensemble-based RoBERTa model and ChatGPT, demonstrate the practical utility of the evaluation framework. For example, our results show that our framework can get an LLM to confidently respond to an extra 20.1\% of low-risk inference tasks that other methods might misclassify as high-risk, and skip 19.8\% of high-risk tasks, which \emph{would} have been answered incorrectly.

\end{abstract}
\begin{document}

\flushbottom
\maketitle
%
%
\thispagestyle{empty}


\section*{Introduction}
\label{sec: introduction}

 Large language models (LLMs) such as OpenAI's GPT \cite{brown2020language, openai2023gpt4} series have emerged as powerful tools capable of diverse natural language processing (NLP) tasks, from simple question-answering to complex narrative generation \cite{wang2022pre, sun2022survey, zhou2023comprehensive}. 
 As these models grow in prominence, concerns about their robustness and reliability, including generalization \cite{hendrycks2021many}, hallucination \cite{lee2023mathematical, ji2023survey}, bias \cite{zhuo2023exploring, ferrara2023should}, and over-confidence \cite{oppenlaender2023mapping}, inevitably arise. Hence, there has been significant interest in evaluating these models' robustness, especially in relation to adversarial attacks \cite{perez2022ignore, subhash2023universal} and for out-of-distribution inputs \cite{pelrine2023towards, moskvichev2023conceptarc}. Research into language models' (LMs) capacity to handle uncertain generation scenarios, especially when the correct answer is not directly provided in the context, have led to the introduction of benchmarks such as SQuAD for generative question answering \cite{rajpurkar2016squad, rajpurkar2018know}. Yet, the challenge of recognizing and mitigating risks in Natural Language Inference (NLI) tasks remains largely uncharted, especially when applying LLMs to applications that require high accuracy and reliability, such as the biomedical and healthcare domains.
 
 Historically, the perceived risk associated with a model's inference and prediction in machine learning was often related to its confidence score \cite{klas2018uncertainty}. Lower confidence in the predicted answer was assumed to signal (relatively) greater risk, implicitly reflecting the model's self-evaluated probability of its correctness. Despite its problems, this conventional approach continues to be used in the deep learning community \cite{loquercio2020general, henne2020benchmarking}, including the transformer-based LLMs \cite{schuster2022confident, vazhentsev2022uncertainty}, owing to its simplicity and computational efficiency (compared to more expensive heuristic methods e.g., those relying on sampling an ensemble of model responses). 

To achieve the confidence level of LLM responses internally, for generative models, various prompts are utilized to infer the reliability of generated answers \cite{si2022prompting, wightman2023strength}. Applications such as chain-of-thought prompting \cite{wei2022chain, wang2022self} are exemplary in employing confidence generated by generative LLMs. Discriminative LLMs, on the other hand, are typically fine-tuned to directly yield softmax probabilities from their final output layer. While the study by \cite{kadavath2022language} confirmed these confidence scores for their calibration capabilities on high-certainty inferences, LLMs have often been noted to produce poorly calibrated confidence scores that don't truly indicate the correctness of the output, as highlighted by \cite{jiang2021can}. Research from \cite{guo2017calibration, jagannatha2020calibrating, jiang2021can, kuhn2022semantic} has aimed to re-calibrate confidence using entropy or by designing binary classifiers based on these scores to detect uncertain inferences. However, a notable gap remains in thorough evaluations of how effectively these confidence-based risk indicators capture the risks LLMs face during inference.

In this study, we argue that a single (often implicit) definition of risk that only relies on over-confidence \cite{hu2021uncertainty, oppenlaender2023mapping}, as has been the case in the majority of research on this topic, may be insufficient. Rather, we advocate for a risk-centric evaluation framework that defines two distinct risk types. We designate these risks as \emph{decision} and \emph{composite risk} and define them formally in the next section. We also present an evaluation framework, including metrics, for objectively measuring these risks. Specific  contributions are as follows:
\begin{enumerate}
    \item We introduce a novel risk-centric framework for better understanding \textit{risk-adjusted inference} in both discriminative and generative LMs. We formalize an LM's decision making as sequential application of a `decision rule' and `selection rule'. Mistakes in each application can lead to two novel kinds of risk, called decision risk and composite risk. 
    
    \item To accommodate these risks, we propose and implement a novel risk-adjusted calibration framework called `Deciding when to Decide' (DwD),  which uses an external decision rule method, compatible with both discriminative and generative LMs. We also propose risk injection functions, applicable to any existing multiple-choice NLI benchmark, for trainin and evaluating DwD. 
    
    \item We present detailed experimental results showing the utility of the framework. We also show that DwD empirically outperforms competitive baselines on several established inference benchmarks by reducing decision and composite risk in underlying LMs by margin of up to 25.3\% and 16.6\%, respectively.
    
    
    \item We present a case study that applies the risk-centric evaluation framework to evaluate the decision and composite risks in a complex real-world inference scenario known as \textit{choice overload}. This case study exemplifies the practical application of the evaluation framework in real-world inference settings, offering insights into its application and utility.
\end{enumerate}

The rest of this paper is structured as follows. Section \ref{related_work}  covers related work on evaluating risk and uncertainty in LMs across various NLP tasks. Next, we define specific risks --decision and composite -- underlying the study in Section \ref{formalism}. Section \ref{DwD} introduces \emph{DwD}, a risk-adjusted calibration implemented as a robust decision rule method compatible with both discriminative and generative LMs. Section \ref{metrics} presents novel evaluation metrics for measuring decision and composite risks, while Section \ref{sec: experiment} provides details on the experimental study, followed by results in Section \ref{results}. Section \ref{casestudy} describes the case study, with the article concluding in Section \ref{conclusion}.

\section*{Related Work}\label{related_work}
Discriminative and generative LLMs have recently achieved impressive performance on multiple inference tasks \cite{bhargava2022commonsense, zhao2023survey}. However, due to documented problems such as hallucination and bias, the research focus is shifting from merely quantifying accuracy to an in-depth, context-sensitive probing of LLMs' robustness, particularly when confronted with risky or uncertain situations \cite{wang2023robustness,shen2023experimental,liang2022holistic}.

The concept of risk has been studied across various disciplines. Historically, risk was perceived as a deviation from the norm, embodying misfortunes and undesirable events, with an underlying assumption of human agency in mitigating adverse outcomes \cite{lupton2013risk}. This perspective  significantly influenced  contemporary discussions on risk, but more recently, attention was focused on formalizing these intuitions by quantifying uncertainty and ambiguity \cite{reddy1996claims, boholm2012semantic}. In practical applications, such as in \textit{Failure Mode and Effects Analysis} (FMEA) \cite{chang2011evaluating, chin2009failure, liu2013risk}, risk evaluation methodologies have been developed to quantify risks based on their probability, severity, and potential for detection, but has faced critiques for its practical limitations in being applied to real-world scenarios. The adoption of quantitative risk metrics also enables the categorization of risks into tiers (low, medium, high) based on the likelihood and consequences of incidents, as seen in various industries \cite{lu2015comprehensive}. In specialized fields, notably healthcare, risk quantification involves additional methodologies, such as calculating the lifetime incidence of diseases within populations \cite{brose2002cancer}. These approaches to defining and evaluating risk underscore the diverse nature of risk management and its significance across different domains, and provide valuable guidance for exploring risk in the context of LMs, as specifically applied to NLI problems.

Systematic methodologies for evaluating LLMs in high-risk scenarios include: `adversarially' attacking LLMs through the use of semantically equivalent adversarial rules \cite{ribeiro2018}, use of adversarial triggers \cite{wallace2019a} and even human-in-the-loop generation of adversarial examples \cite{wallace2019}. Recent studies also explore the riskiness of certain LLM prompts \cite{halawi2023overthinking, shi2023large, turpin2023language}. While such research spotlights LLMs' vulnerabilities when adversarially prompted, we still expect LLMs to make judicious decisions about navigating a risk-reward tradeoff by being self-aware of when they might have a higher probability of going wrong. Besides, although `risk' is frequently mentioned in adversarial attack research, as also in related literature on assessments of safety and bias \cite{ferrara2023should, zhou2022towards}, robustness-accuracy characteristics \cite{ko2023robustness}, and out-of-distribution (OOD) generalization \cite{grangier2022trade, xu2021raise}, a formalism is still lacking that identifies and defines multiple types of risk.  This paper proposes such a general formalism, including identifying two different risk categories that are relevant to LLMs, and novel metrics for measuring risk.

Initial research on LLMs' `self-understanding' of their own uncertainty, especially in the deep learning literature, has predominantly relied on interpreting raw softmax probabilities of the final output layer as `confidence' scores \cite{vasudevan2019towards}. 
While studies such as \cite{guo2017calibration} have flagged these scores as potentially misleading and not \emph{genuinely} capturing the model's true uncertainty, The study in \cite{kadavath2022language} highlighted that generative LLMs exhibit commendable calibration properties facing certain situations. These models can accurately predict which questions they will be able to answer correctly on diverse NLI tasks based on their confidence scores. Nevertheless, a quantitative assessment of risk, even for such models, has been lacking.

Jiang et al. \cite{jiang2021can} suggest that, when faced with uncertain situations,
LLMs can sometimes be poorly calibrated, with the confidence score estimation barely being correlated with the likelihood of the output being correct. Building on these observations, studies such as \cite{jagannatha2020calibrating, jiang2021can, kuhn2022semantic} have endeavored to re-calibrate confidence, using mechanisms like entropy, or by crafting binary classifiers based on the given confidence scores. More recently, some authors \cite{zhou2023navigating} have focused on generative models, probing linguistic hedge markers in the models' outputs to evaluate their ability to discern uncertain situations. Research, such as by \cite{rajpurkar2016squad, rajpurkar2018know}, has also emphasized tasks such as generative question answering, particularly on information extraction problems. The authors set up unanswerable inference tasks wherein related tokens are present in the context but do not constitute the correct answer to the given questions. In this scenario, researchers can evaluate whether the surface similarity (such as overlap) between context and candidate answer is the main distractor for LMs to predict if the instance is answerable or not. A further strand of research, such as by \cite{yin-etal-2023-large}, proposes benchmarks aiming to spotlight areas of knowledge where LLMs grapple with uncertainty. Concurrently, efforts in \cite{collins2023human} seek to minimize uncertainty in human-AI contexts by addressing risks originating from human errors.

In our work, we propose a novel re-adjusted confidence calibration for both generative and discriminative LLMs, aimed at discerning uncertain situations. We utilize a binary classifier calibration, previously shown to have exceptional performance \cite{kamath2020selective,jiang2021can}, as an experimental baseline. Rather than presenting a new benchmark, we devise a suite of risk injection functions designed to emulate `unanswerable' uncertain inference scenarios, and that can be applied to any existing multi-choice NLI benchmarks. 

\section*{Decision and Composite Risk: Definition and Evaluation Framework} \label{formalism}
For terminological convenience, let us denote a multi-choice inference instance as $i=(q, Y)$ in an NLI benchmark $I$ (consisting of a set of such instances), where $q$ is a \emph{prompt} (which can be, but is not necessarily limited to being, a proper `question') and $Y$ is a set of candidate choices. Both discriminative and generative LMs aim to identify the correct choice:

\begin{equation}
\centering
\hat{y}=\underset{y \in Y}{\mathrm{argmax}}\ P_{LM}(y|q),
\label{equ1}
\end{equation} 

Here, $P_{LM}(y|q)$ refers to the probability assigned by the LM to each choice $y$, indicating its likelihood of being correct. This probability is effectively estimated by a confidence score $c$ assigned to each choice $y \in Y$, regardless of the actual presence of a correct answer within $Y$. Instances are termed \emph{ambiguous} when lacking a `ground-truth' correct answer $\hat{y} \in Y$, with an indicator $\hat{i}$ employed for clarity: $\hat{i}=1$ and $\hat{i}=0$ denote the presence and absence of a ground-truth answer, respectively.

Considering inference instances, we can model decision-making in LMs as two sequential application: of the decision rule $dr$, followed by the selection rule $sr$. We reformulate Eq. (3.1) as follows:

\begin{equation}
\hat{y}= sr(q, Y) [P_{LM}(y|q) \cdot \mathbbm{1}({dr(q, Y) = 1})],
\end{equation} 

Here, $sr(q, Y) = \underset{y \in Y}{\mathrm{argmax}}\ y $. We designate a \textit{selective NLI (sNLI)} system as a base LM that incorporates both a decision rule and a selection rule. The decision rule operates as a binary classifier: when $dr(q, Y)=1$, the model responds; when $dr(q, Y)=0$, it abstains. Discriminative LMs, not explicitly equipped with a $dr$, default to $dr(q, Y)=1$ and hence, always attempt to make a prediction, which simplifies Eq. (3.2) back to the original form Eq. (3.1). Contrasting sharply with the all-responsive nature of discriminative models, generative LMs (internally) employ a more advanced $dr$, and may choose not to respond for certain statements with outputs like ``I don't understand'' or ``None of the answers seem to be correct'', implying $dr(q, Y) = 0$. 

When $dr(q, Y)=1$, the \emph{selection rule} $sr$ is invoked. Commonly, and as assumed here, $sr$ predicts $\hat{y}'$ with the highest $P_{LM}(y|q)$, which is estimated by confidence ($c \in C$), with ties broken arbitrarily. Alternative selection rules can be devised, but are rare, and not considered herein.

Using this terminology, we can define decision risk as follows:
\begin{definition}[Decision Risk] 
\label{def: decisionRisk}
Given an instance $i=(q,Y)$ and confidence set $C$ over $Y$, the \emph{decision risk} $r_d$ is set to 1 (and is otherwise 0) \emph{iff} at least one of two conditions is met: (1) the instance is unambiguous ($\hat{i} =1$) but $dr(q, Y) = 0$; and (2) the instance is ambiguous ($\hat{i} = 0$) but $dr(q, Y) = 1$.
\end{definition}

\begin{figure*}[ht]
  \centering
  \includegraphics[width=0.7\textwidth]{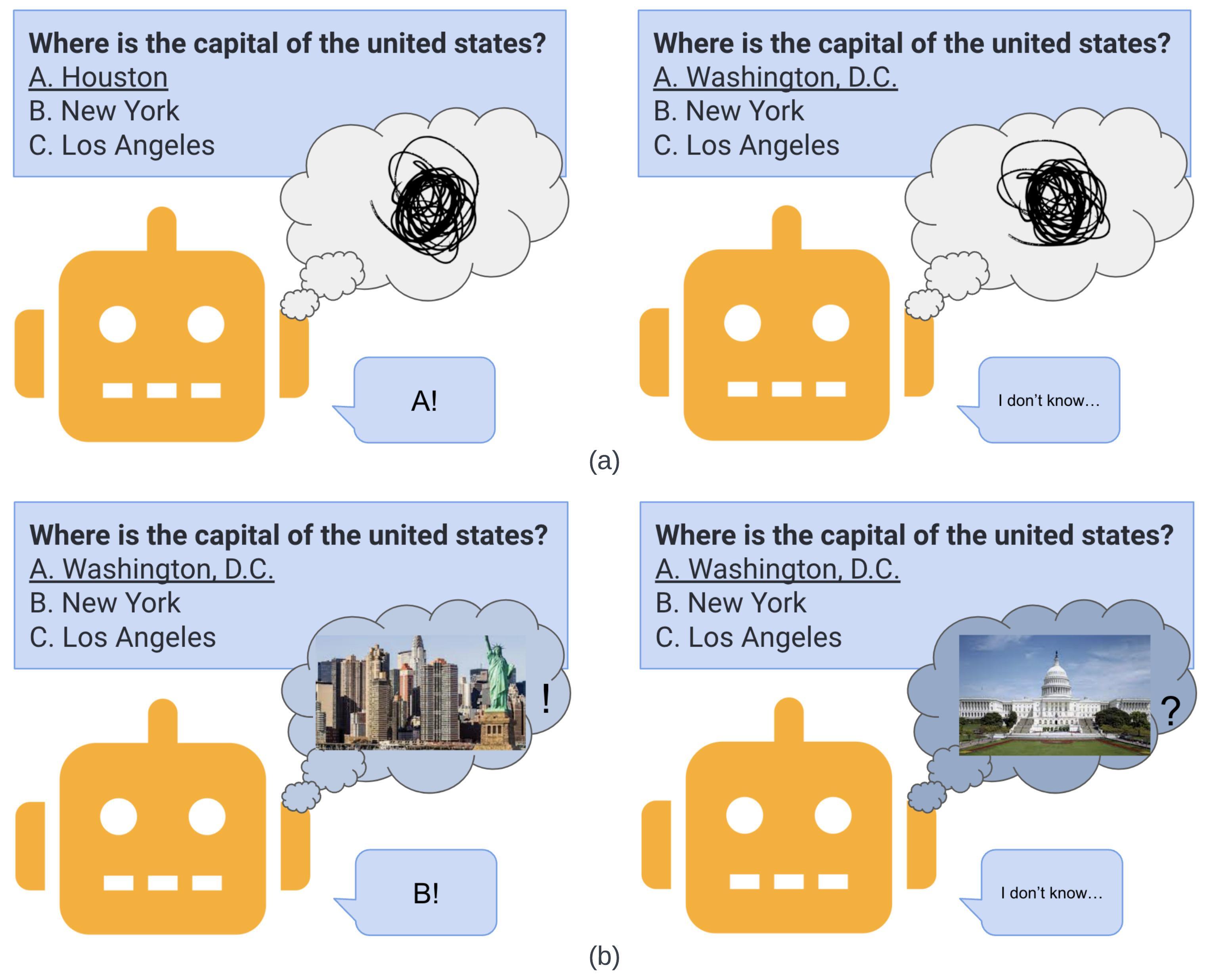}
  \caption{Illustration of (a) decision risk and (b) composite risk (b) in LMs in NLI tasks.}
  \label{fig:risk_illustration}
\end{figure*}

Figure \ref{fig:risk_illustration} (a) provides illustrative examples of decision risk occurrence in LMs. In the first scenario (left), the LM confronts an ambiguous inference instance, where there is a prompt but the candidate choices do not contain a ground-truth answer. Ideally, the LM's decision-rule should choose not to provide an answer. Contrary to this expectation, the LM erroneously commits to an incorrect option. The second scenario (right) shows a situation where, despite a correct answer being available among the candidate choices, the decision rule opts to withhold a response, signaling ``I don't know'', potentially due to a lack of confidence in its predictive accuracy. Both instances highlight the emergence of decision risk during natural language inference can emerge, either when LMs inappropriately respond or when they unnecessarily abstain.

Decision risk can be an important concern in many human-facing and otherwise critical applications, such as clinical decision-making \cite{minutolo2016fuzzy, popat2023embracing}. In such domains, ambiguity is not uncommon for a variety of complex reasons, including expert disagreement, and evolving situational knowledge. Note that, in some applications, one of the two conditions in the definition might be more consequential than the other; however, in this paper, we treat both equally as decision risks.

Next, we define the \emph{composite risk}, which is motivated by the common situation where all instances are technically unambiguous but where an underlying LM finds some instances riskier (more likely to get wrong) than others. Even in the absence of ambiguity, composite risk can be used to quantify both potential under-confidence and over-confidence in the model's performance. 

\begin{definition}[Composite Risk] 
\label{def: compositeRisk}
Given an instance $i = (q, Y)$ with a unique correct answer $\hat{y} \in Y$ and a confidence set $C$ over $Y$, the \emph{composite risk} $r_s$ is set to 1 (and is otherwise 0) \emph{iff} at least one of two conditions is met: (1) $dr(q, Y)=1$ but the selection rule makes an incorrect prediction $\hat{y}'\neq \hat{y}$; and (2) $dr(q, Y)=0$ but the selection rule of the model yields a correct prediction $\hat{y}'=\hat{y}$.
\end{definition}

Figure \ref{fig:risk_illustration} (b) illustrates intuitive examples of composite risk by concurrently examining the outcomes of decision and selection rules in risk evaluation. The figure and definition clarify that composite risk first appears when the selection rule erroneously identifies an incorrect answer, when the decision rule chooses to answer. In an ideal model, the decision rule would choose not to answer when it is less certain about the correctness of its prediction. The second example of composite risk is seen when the selection rule accurately identifies the correct answer, but the decision rule withholds this correct response, choosing instead to respond with ``I don't know.'' These examples highlight the interaction between decision and selection rules within the evaluation framework of composite risk.

In prior work, the (empirical) risk is formally characterized as $\mathbb{E}[\ell(\hat{y}', y) \cdot dr(q, Y)]$, where $\ell$ typically represents a 0/1 error loss function \cite{geifman2017selective}. This offers a streamlined representation of the composite error while preserving its original semantics. According to \cite{guo2017calibration}, when the confidence of the predicted answer $\hat{y}'$ converges to $1/K$, with $K$ being the cardinality of the candidate answer set $Y$, uncertainty is at its peak. According to this study, such a scenario also maximizes the likelihood of composite risk.

\section*{Risk-adjusted Calibration Approach}\label{DwD}

As previously noted, the discriminative models' decision rule was inherently designed to respond to every query, defaulting to $dr(q,Y)=1$. Modern generative models, such as ChatGPT, have not disclosed their decision-making protocols, making any decision rules within them unpredictable from a user's standpoint. To evaluate and mitigate the decision and composite risk across both discriminative and generative models, an external decision rule method that is compatible with both types of models, and that is somewhat independent of the LM itself (and hence, generalizable), is clearly motivated. An ideal decision rule should use all available information, such as the instance prompts, as well as the confidence outputs from the underlying LLM, to minimize the risks defined earlier. 

Previous studies \cite{guo2017calibration, vazhentsev2022uncertainty} have developed fundamental techniques for the \emph{re-calibration} of confidence scores, aimed at more precisely capturing a model's intrinsic uncertainty. We expand on this idea significantly by proposing a novel \emph{risk-adjusted} calibration method called `Deciding when to decide' or DwD. An external \emph{decision rule} method in the architecture helps minimize the decision and composite risks of LMs, especially in high-risk inference scenarios. Notably, DwD does not have strong dependencies on the underlying LMs, enabling it to be suitable for a variety of models without knowing their internal workings. As we subsequently demonstrate in experiments, this feature makes it well-suited for risk-adjusted inference even for blaxk-box commercial models like OpenAI's GPT-4.

Similar to its predecessors, the decision rule in DwD operates as a binary classifier, utilizing traditional machine learning techniques. However, it distinguishes itself from previous work by tackling two principal challenges: the injection of risk into the training process, and calibration refinement with risk adjustment. Concerning the former, DwD injects risk into training set construction, a step not explicitly considered by earlier calibrators. Existing NLI benchmarks usually provide a definitive answer for all instances. Theoretically, an effective decision rule could use this knowledge and just answer all instances to minimize the decision risk. Here, to introduce risk-injected instances in the training of DwD approach, we introduce \emph{Risk Injection Functions} (RIFs) which can effectively turn an instance $i=(q, Y)$ with an unambiguous correct answer into an `ambiguous' instance $i'$, with no correct answer:

\begin{itemize}
	\item {\bf Wrong-Question RIF (WQ):} Retain the candidate choice set $Y$  but replace the original prompt $q$ with a new prompt $q'$ from an unrelated instance in the same benchmark; 
	\item {\bf No-Right-Answer RIF (NRA):} Retain the prompt $q$ and all \emph{incorrect} choices $Y-\hat{y}$ in the new candidate choice set $Y'$; also, add to $Y'$ a choice from another unrelated instance in the same benchmark.
\end{itemize}

These RIFs expose DwD to diverse risk scenarios, ensuring that the method is well-equipped to handle real-world challenges.

DwD also refines its calibration by leveraging a comprehensive feature set including prompt length (in terms of the number of characters), the length of the predicted answer generated by the LLM, the confidence score of each candidate answer, the standard deviation of confidence scores across choices, the sentence embedding (which we obtained using the pre-trained \emph{nq-distilbert-base-v1} model in \emph{SentenceTransformer} \cite{reimers2019sentence}), the  similarity between the prompt and each candidate choice, and the standard deviation of these embedding similarities. This diverse array of features is processed through a random forest classifier, trained on an equal mix of original and risk-injected instances perturbed by one of the RIFs, to predict the likelihood of an instance being labeled positively by DwD. By using a diverse set of features, our aim is to help DwD be as robust as possible so that it is able to navigate decision and composite risk better, and utilizes RIF training to maximum advantage. With these features, the \emph{DwD} method is trained using a random forest classifier with equal numbers of original and risk-injected instances (following the application of one RIF) as the training set. The probability of an inference instance being labeled as positive is interpreted as the confidence of DwD asserting $dr(q, Y)=1$, which is used to estimate both the decision and composite risks. Note that the \emph{DwD} approach is specifically trained on a discriminative LLM's confidence distribution, as generative LLMs such as GPT-3.5-turbo can only provide a `fuzzy' confidence estimate. In \emph{Results}, we show that, although the approach was exclusively trained on a discriminative LLM's confidence distribution, it is adaptable to generative LLMs.

\section*{Evaluation Metrics}\label{metrics}

\subsection*{Decision Risk}

For the decision risk evaluation to be non-trivial, it is essential to include ambiguous inference instances—those without a correct option within the candidate choice setting in the evaluation sets. We employ the WQ and NRA RIFs introduced before to perturb the original instances within the existing NLI benchmark's evaluation set. The decision risk evaluation-dataset will thus consist of a balanced mix of both original and risk-injected instances. Note that there is a deliberate separation between the instances perturbed for training purposes and those adjusted for evaluation to ensure no overlap.

Since an effective decision rule $dr(q, Y)$ over LLMs' confidence score should return 0 for risk-injected instances and 1 for risk-free instances to minimize decision risk, we can evaluate the efficacy of $dr$ by calculating the inverse proportion of the decision risk,
\begin{equation}
    P = P(dr(q, Y) = \hat{i}|i\in \{I \cup f(I)\}, dr)
\end{equation}

 when a certain RIF $f$ was applied to generate $i'$ in the evaluation set. Without loss of generality, we evaluate the robustness of $dr$ in two different scenarios: \emph{In-domain (ID)} and \emph{Out-of-domain (OOD)}. In an ID evaluation, the same RIF applied for training a decision rule is re-applied to an evaluation set containing unseen instances. Hence, the decision rule method has an opportunity to `learn' the RIF from the training data. In contrast, the OOD evaluation, meant to be harder and more realistic, uses different RIF(s) than the ones used during training; hence, the decision rule has no knowledge of these RIFs during the learning phase. These intuitions are illustrated in Figure \ref{fig:framework-details}.

\subsection*{Composite Risk}

Unlike decision risk, which solely focuses on the decision risk, the composite risk depends on both the decision risk and selection risk. To capture the complexity, we first use two metrics, \emph{risk specificity} and \emph{risk sensitivity}, inspired by epidemiological practices \cite{runyan1998prevalence}: 
\begin{equation}
    P_{spe} = P(dr(q, Y)=0|sr(q, Y)\neq \hat{y})
\end{equation}
\begin{equation}
    P_{sen} = P(dr(q, Y)=1|sr(q, Y)= \hat{y})
\end{equation}

Risk specificity evaluates the performance of sNLI systems when the selection rule leads to an incorrect prediction, whereas risk sensitivity measures system performance when the selection rule can accurately identify the correct answer. A low value in either of the two indicates a significant composite risk; for instance, overly aggressive decisions relative to the accuracy of the selection rule yield low risk specificity.

\begin{figure*}[t]
  \centering
  \includegraphics[width=\textwidth]{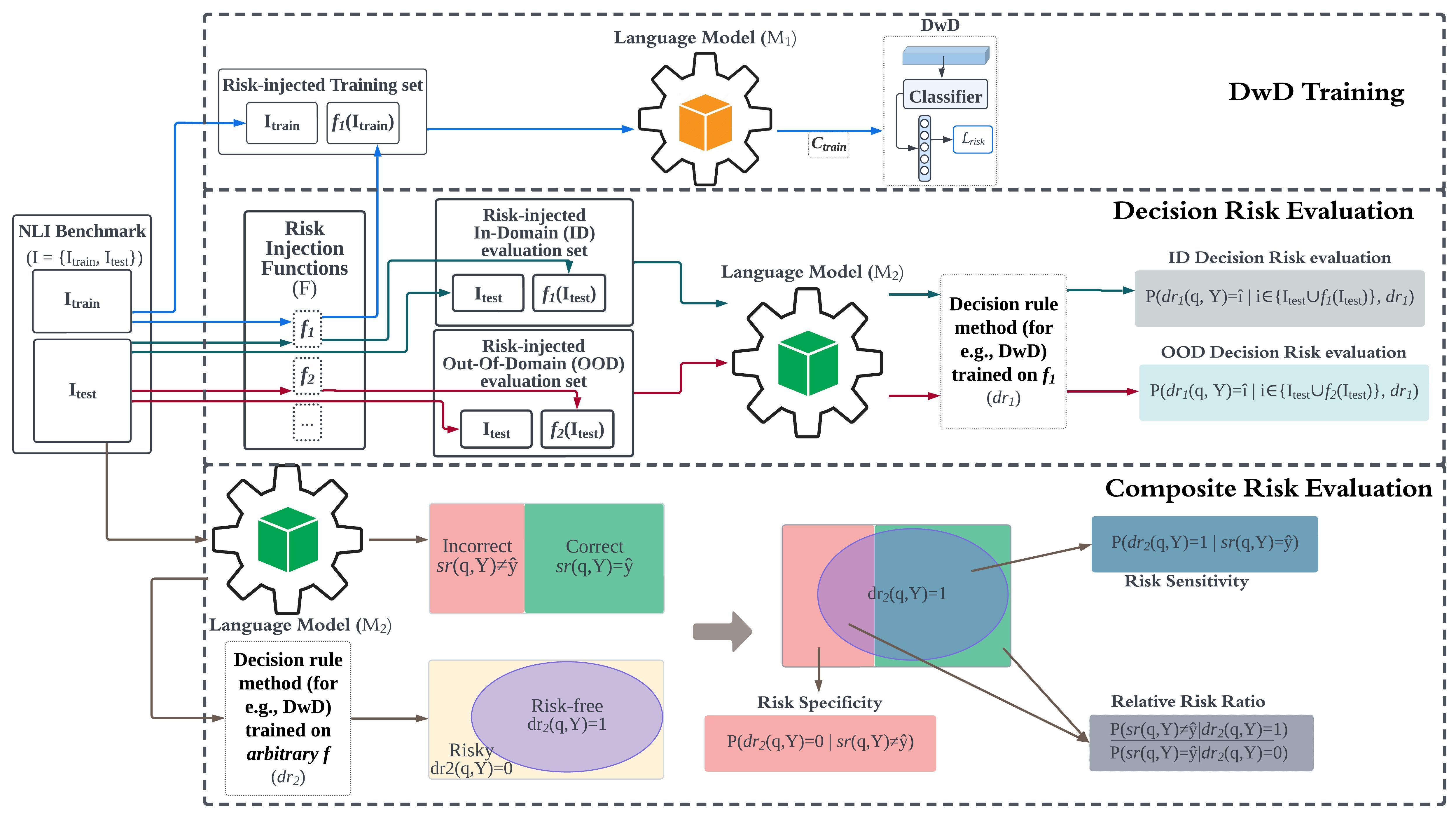}
  \caption{The risk-centric framework for evaluating LLMs on NLI tasks. Symbols used in the figure are further described in the main text.}
  \label{fig:framework-details}
\end{figure*}

To describe the trade-off between answering more questions and decreasing composite risks, we also use the \emph{relative risk ratio} (RRR) metric \cite{tenny2022relative}:
\begin{equation}
    RRR = \frac{P(sr(q, Y)\neq \hat{y}|dr(q, Y)=1)}{P(sr(q, Y)= \hat{y}|dr(q, Y)=0)}
\end{equation}
where the probability condition is reversed compared to risk sensitivity and specificity, defined earlier. When $RRR$ is significantly smaller than 1 (e.g., at the 95\% confidence level), it implies that $dr$ significantly reduces composite risk when it decides to answer ($dr=1$), compared to when it abstains. Figure \ref{fig:framework-details} presents a full overview of the evaluation framework for decision and composite risk.

\section*{Experimental Study}
\label{sec: experiment}
\paragraph{Datasets}
We use four established NLI benchmarks (aNLI \cite{anli}, HellaSwag \cite{hellaswag}, PIQA \cite{piqa}, and SocialIQA \cite{socialiqa}), modeled as multiple-choice tasks with one correct answer per prompt. The decision risk evaluation is conducted on a balanced evaluation set that comprises an equal number of original inference instances (sourced from the evaluation sets of each benchmark) and the corresponding perturbed instances. These perturbed instances are generated by applying one RIF and are matched in size with the original instances. Conversely, the evaluation of composite risks is solely performed on the original instances within each benchmark's evaluation dataset. To explore the risk profiles of both discriminative and generative language models, we use two prominent models: RoBERTa-large Ensemble \cite{robertaensemble} and OpenAI's GPT-3.5-Turbo.

\paragraph{Decision Rule}

Preliminary evaluation of GPT-3.5-Turbo shows that its aggressive built-in decision rule chose to respond to 88.7\% of perturbed, high-risk ambiguous instances. This rule yielded a decision risk accuracy of 55.7\%, which is only marginally better than the random baseline. Considering these findings, our experiments implement external decision rule methods for sNLI systems to effectively navigate and improve upon the inherent limitations of the built-in decision rule.

Except for DwD, three decision rule baseline methods are employed in the experiments. Like DwD, all baselines described below treat the LMs as black boxes and do not require access to the model's internal representations. Rather, for each instance, they only need the output (the confidence set $C$) of the model.

{\bf Random Baseline.} Given an instance (into which risk may or may not have been injected), this baseline ignores the confidence set $C$ and randomly chooses between 1 (risk-free) and 0 (risky) with equal probability. Hence, if risk has been injected in half the instances in the evaluation set, the accuracy of its prediction will always be 0.5 (in expectation), and is intended to serve as a useful reference for evaluating more advanced decision rule methods.

{\bf ConfStd Baseline.} Inspired by \emph{MaxProb} \cite{hendrycks2016baseline}, this baseline uses the \emph{standard deviation} among all candidate choices' confidences (referred to as \emph{ConfStd}) to make its decision. Using MaxProb directly was found to yield near-random performance when evaluating decision risk (see \textit{Supplementary Information}). A lower \emph{ConfStd} corresponds to a higher risk of answering incorrectly. We use the training set of a benchmark to determine the optimal \emph{ConfStd} threshold below which a decision of 0 (risky) would be returned. This threshold is then applied during evaluation. 

{\bf Calibrator Baseline.} Finally, we used another baseline that relies on training a \emph{calibrator} \cite{dong2018confidence, kamath2020selective}. We opted to use the random forest model as the binary classifier, aligning with the consistency of our proposed \emph{DwD} method. Inspired by the experimental settings in \cite{kamath2020selective}, we designed the calibrator baseline using the prompt length (in terms of the number of characters), predicted answer ($\hat{y}$) length, and each candidate choice's confidence, as features.

For the purposes of training decision rule baselines (excluding the Random Baseline) and DwD approach, risk-injected instances are labeled with 0, and original instances with 1. For conducting in-domain (ID) evaluations, we use each of the RIFs in turn during both training and evaluation, and report results separately for each RIF. When conducting out-of-domain (OOD) evaluations, however, another set of output is obtained for each RIF utilized during training, based on the application of different RIFs in the evaluation phase.

\paragraph{Selection Rule}
We consider a standard confidence-based selection rule in this paper: given the confidence set $C$, we select $\hat{y}'$ as the selected choice, where $\hat{y}'$ is the choice assigned the highest confidence in $C$. Ties are broken arbitrarily.
\section*{Results}\label{results}
\paragraph{Evaluating ID and OOD Decision Risk}
Table \ref{tab: IDandOOD} reports the accuracy results (equivalent to 1$-$the proportion of decision risks) for \emph{RoBERTa Ensemble} incorporating various decision rule methods (DwD and the three baselines) on in-domain (ID) and out-of-domain (OOD) decision risk evaluation datasets. In evaluating ID decision risk, across all benchmarks and settings, the accuracy of \emph{RoBERTa Ensemble}, when guided by the proposed DwD method, outperformed its accuracy in conjunction with ConfStd and Calibrator, by significant margins often exceeding 20 percent. This suggests the efficacy of learning-based methods (such as DwD and Calibrator), and of RIF-based training, in particular. In contrast, ConfStd, which directly utilizes RoBERTa's confidence as a decision rule, showed near-random performance for the PIQA benchmark, with some improvements on other benchmarks. This finding aligns with the observation that the raw confidences of LLMs themselves tend to be poorly calibrated when confronted with ambiguity and uncertainty \cite{jiang2021can}.

As the nature of decision risks confronted by LLMs is typically unknown, handling OOD decision risk is expected to be harder than handling ID risk. Table \ref{tab: IDandOOD} (b) illustrates that \emph{RoBERTa Ensemble}, utilizing a DwD-generated decision rule trained with either WQ or NRA function, achieved the best performance with an average accuracy typically above 60 percent, and in some cases, above 75 percent, outperforming other baselines by substantial margins. On average, DwD had a decline of 9 percent compared to its ID performance, and the absolute estimates suggest that \emph{RoBERTa Ensemble} can be properly re-calibrated when exposed to both ID and OOD decision risks, when coupled with the correct decision rule method. The OOD results also confirm that RoBERTa's raw confidence cannot directly be used as a reliable decision rule, but the DwD-adjusted decision rule can predict decision risk effectively.

\begin{table}[htbp]
\centering
    \begin{subtable}
        \centering
        \begin{tabular}{lllll}
\hline
              & \textbf{aNLI} & \textbf{HellaSwag} & \textbf{PIQA} & \textbf{SocialIQA} \\ \hline
ConfStd\text{\scriptsize$_{WQ}$}      & 0.640**       & 0.610**            & 0.550**       & 0.600**            \\
ConfStd\text{\scriptsize$_{NRA}$}     & 0.460         & 0.630**            & 0.410         & 0.550**            \\ \hline
Calibrator\text{\scriptsize$_{WQ}$}   & 0.630**       & 0.730**            & 0.560**       & 0.620*             \\
Calibrator\text{\scriptsize$_{NRA}$}  & 0.510         & 0.660**            & 0.540*        & 0.500              \\ \hline
DwD\text{\scriptsize$_{WQ}$}          & \textbf{0.940**}       & \textbf{0.940**}            & 0.830**       & \textbf{0.790**}            \\
DwD\text{\scriptsize$_{NRA}$}        & 0.830**       & 0.820**            & \textbf{0.840**}       & 0.630**            \\ \hline
\end{tabular}
       \caption*{(a)}
       \label{tab:ID}
    \end{subtable}
    \begin{subtable}
        \centering
        \begin{tabular}{lllll}
\hline
              & \textbf{aNLI} & \textbf{HellaSwag} & \textbf{PIQA} & \textbf{SocialIQA} \\ \hline
ConfStd\text{\scriptsize$_{WQ}$}      & 0.535         & 0.605**            & 0.475         & 0.580**            \\
ConfStd\text{\scriptsize$_{NRA}$}     & 0.630**       & 0.570**            & 0.540**       & 0.655**            \\ \hline
Calibrator\text{\scriptsize$_{WQ}$}   & 0.520         & 0.685              & 0.535         & 0.580*             \\
Calibrator\text{\scriptsize$_{NRA}$} & 0.520         & 0.645**            & 0.505         & 0.570**            \\ \hline
DwD\text{\scriptsize$_{WQ}$}          & 0.770**       & 0.750**            & \textbf{0.680**}       & 0.620*             \\
DwD\text{\scriptsize$_{NRA}$}        & \textbf{0.885**}       & \textbf{0.900**}            & 0.630**       & \textbf{0.665**}            \\ \hline
\end{tabular}
        \caption*{(b)}
        \label{tab:OOD}
     \end{subtable}
     \caption{Accuracy of \emph{RoBERTa Ensemble} using a decision rule (ConfStd / Calibrator / DwD) when risk-injected instances are present. The performance is reported for both in-domain (a) / out-of-domain (b) scenarios. The best performance is shown in bold. *, ** indicate statistical significance with respect to \emph{Random} baseline (which consistently scores 0.5) at the 90\%, 95\% confidence level, respectively.}
     \label{tab: IDandOOD}
\end{table}

\paragraph{Evaluating Composite Risk}

Table \ref{table: sensitivity} and \ref{table: specificity} show the results for composite risks. The \emph{RoBERTa Ensemble} using the WQ-trained DwD decision rule achieved the highest sensitivity and specificity across all benchmarks. Both ChatGPT and \emph{RoBERTa Ensemble} tend to yield confidence distribution with high `reference' values for `risk-free' instances. With the DwD rule, over 90\% of instances in \emph{aNLI} and \emph{HellaSwag} benchmarks were accurately answered, significantly reducing composite risk. In instances where LLMs might err if left unguided, DwD-guided \emph{RoBERTa Ensemble} maintains the best performance. However, it is less specific than sensitive, recording an average specificity of 0.573 over four benchmarks.

\begin{table}[ht!]
\centering
\begin{tabular}{llllll}
\hline  
&                 & \textbf{aNLI$\uparrow$}                         & \textbf{HellaSwag$\uparrow$}                    & \textbf{PIQA$\uparrow$}                         & \textbf{SocialIQA$\uparrow$}                  \\ \hline
\multirow{6}{*}{RoBERTa} & ConfStd$_{WQ}$ & 0.801                    & 0.787                         & 0.591                    & 0.681                         \\
& ConfStd$_{\_NRA}$     & 0.733                    & 0.784                         & 0.485                    & 0.643                         \\ \cline{2-6} 
& Calibrator$_{\_WQ}$   & 0.645                    & 0.787                         & 0.566                    & 0.630                         \\
& Calibrator$_{\_NRA}$  & 0.541                    & 0.691                         & 0.531                    & 0.537                         \\\cline{2-6} 
& DwD$_{\_WQ}$  & \textbf{0.928}           & \textbf{0.959}                & \textbf{0.836}           & \textbf{0.781}                \\
& DwD$_{\_NRA}$ & 0.818                    & 0.843                         & 0.811                    & 0.599                         \\ \hline
\multirow{2}{*}{GPT-3.5-Turbo} & DwD$_{WQ}$         & 0.912 & 0.954 & 0.826  & 0.758 \\
                         & DwD$_{NRA}$        & 0.846      & 0.727   & 0.824          & 0.516\\

\hline

\end{tabular}
\caption{Sensitivity of all sNLI systems (selective prediction methods + base language model RoBERTa) on four original NLI benchmarks. We use one unified language model (RoBERTa) in all sNLI systems to better compare the effects of different selective prediction methods in the sNLI systems. The first column represents the selective prediction methods used in sNLI systems. The subscripts of values in the first row indicate the data augmentation function used in the training process. The best performances are marked in bold.}
\label{table: sensitivity}
\end{table}

\begin{table}[ht!]
\centering
\begin{tabular}{llllll}
\hline  
&                 & \textbf{aNLI$\uparrow$}                         & \textbf{HellaSwag$\uparrow$}                    & \textbf{PIQA$\uparrow$}                         & \textbf{SocialIQA$\uparrow$}                  \\ \hline
\multirow{6}{*}{RoBERTa} & ConfStd$_{WQ}$ & 0.477                    & 0.232                         & 0.270                    & 0.315                         \\
& ConfStd$_{\_NRA}$     & 0.401                    & 0.228                         & 0.190                    & 0.258                         \\ \cline{2-6} 
& Calibrator$_{\_WQ}$   & 0.322                    & 0.381                         & 0.443                    & 0.446                         \\
& Calibrator$_{\_NRA}$  & 0.500                    & 0.338                         & 0.325                    & 0.394                         \\\cline{2-6} 
& DwD$_{\_WQ}$  & \textbf{0.580}           & \textbf{0.684}                & \textbf{0.576}           & \textbf{0.531}                \\
& DwD$_{\_NRA}$ & 0.560                    & 0.569                         & 0.568                    & 0.519                         \\ \hline
\multirow{2}{*}{GPT-3.5-Turbo} & DwD$_{WQ}$         & 0.120 & 0.054 & 0.206  & 0.231 \\
                         & DwD$_{NRA}$        & 0.160      & 0.170   & 0.188          & 0.444\\

\hline

\end{tabular}
\caption{Specificity of all sNLI systems (selective prediction methods + base language model RoBERTa) on four original NLI benchmarks. We use one unified language model (RoBERTa) in all sNLI systems to better compare the effects of different selective prediction methods in the sNLI systems. The first column represents the selective prediction methods used in sNLI systems. The subscripts of values in the first row indicate the data augmentation function used in the training process of corresponding selective prediction methods. The best performances are marked in bold.}
\label{table: specificity}
\end{table}

\begin{table}[ht!]
\centering
\begin{tabular}{llllll}
\hline  
&                 & \textbf{aNLI$\downarrow$}                         & \textbf{HellaSwag$\downarrow$}                    & \textbf{PIQA$\downarrow$}                         & \textbf{SocialIQA$\downarrow$}                  \\ \hline
\multirow{6}{*}{RoBERTa} & ConfStd$_{WQ}$ & 0.242                    & 0.195                         & 0.348                    & 0.325                         \\
& ConfStd$_{\_NRA}$     & 0.247                    & 0.191                         & 0.348                    & 0.335                         \\ \cline{2-6} 
& Calibrator$_{\_WQ}$   & 0.268                    & 0.224                         & 0.351                    & 0.275                         \\
& Calibrator$_{\_NRA}$  & 0.251                    & 0.197                         & 0.346                    & 0.265                         \\\cline{2-6} 
& DwD$_{\_WQ}$  & \textbf{0.237}           & \textbf{0.182}                & 0.308           & \textbf{0.257}                \\
& DwD$_{\_NRA}$ & 0.247                    & 0.193                         & 0.342                    & 0.279                         \\ \hline
\multirow{2}{*}{GPT-3.5-Turbo} & DwD$_{WQ}$         & 0.503 & 0.560 & \textbf{0.227}  & 0.405 \\
                         & DwD$_{NRA}$        & 0.248      & 0.203   & 0.342          & 0.273\\

\hline

\end{tabular}
\caption{Relative risk ratios (RRRs) of all sNLI systems (selective prediction methods + base language model RoBERTa) on four original NLI benchmarks. We use one unified language model (RoBERTa) in all sNLI systems to better compare the effects of different selective prediction methods in the sNLI systems. The first column represents the selective prediction methods used in sNLI systems. The subscripts of values in the first row indicate the data augmentation function used in the training process of corresponding selective prediction methods. The best performances are marked in bold.}
\label{table: rrr}
\end{table}

Table \ref{table: rrr} shows that with various decision rules, both \emph{RoBERTa Ensemble} and ChatGPT significantly reduced composite risks ($RRR<1$ at 95\% confidence). Again, DwD calibration (trained using \emph{WQ} RIF) obtained the lowest overall RRR.  The average RRR of 0.246 indicates that the risk ratio when DwD prompts RoBERTa to make a prediction is around 25\% compared to the risk ratio when DwD prompts RoBERTa to skip the instances. Interestingly, using the DwD method (trained with WQ RIF) as a decision rule, ChatGPT outperforms RoBERTa on PIQA. Despite DwD being exclusively trained on RoBERTa's confidence distribution, it generalizes well, potentially guiding other LLMs like ChatGPT in confidently answering risk-free inferences. This is notable because models such as RoBERTa, unlike some of the more recent black-box LLMs, are more freely available in the open-source community and require fewer computational resources to fine-tune.

\paragraph{Visualizing Risk-Coverage Tradeoff of DwD}

\begin{figure}[t]
  \centering
  \includegraphics[width=0.5\textwidth]{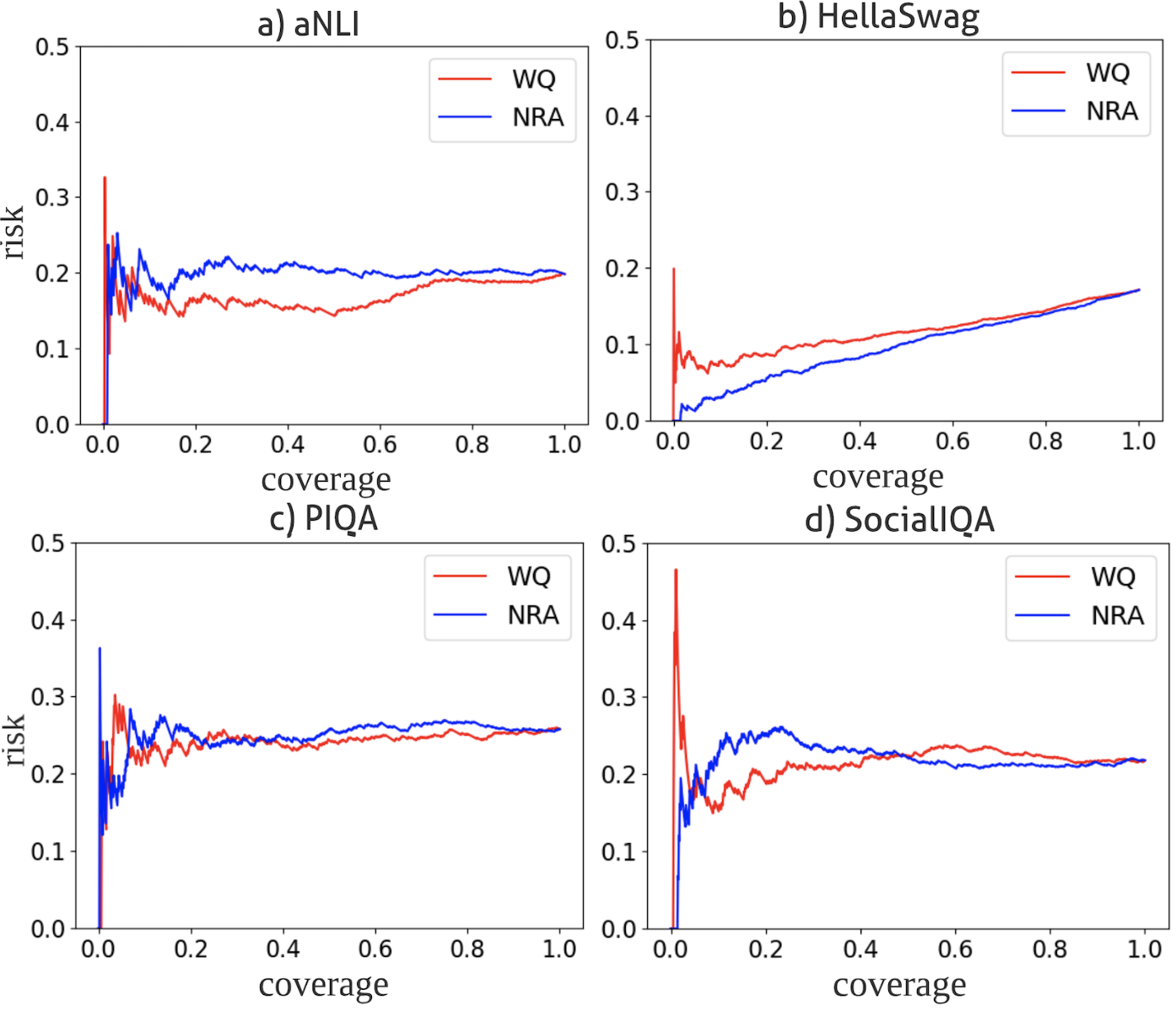}
  \caption{Risk-coverage curves for RoBERTa ensemble model that uses the proposed DwD method (WQ- and NRA-trained versions) as a decision rule on all four benchmarks.}
  \label{fig:risk-coverage}
\end{figure}

We briefly discuss the risk-coverage tradeoff of RoBERTa by plotting the risk-coverage curves (in Fig \ref{fig:risk-coverage}) for each benchmark, when the model was incorporated with the \emph{DwD} decision rule. We expect a robust LLM with an effective risk-adjusted calibration to exhibit a lower aggregate risk when dealing with instances deemed as least risky, compared to riskier instances. Crucially, these risk-coverage curves suggest that the proposed decision rule can drastically decrease inference error across all four benchmarks. For example, with HellaSwag at 90\% coverage, the DwD calibration prompts the RoBERTa system to abstain from 1.8\% of incorrect instances, rising to 3.3\% at 85\% coverage. Considering the LLM's near-human performance, this improvement can significantly impact real-world applications, particularly given the cost-effectiveness of decision rule training. Results for NQ are not shown as it was found to be overfitting when evaluating composite risks, as discussed earlier, and its risk-coverage tradeoff is hence trivial.

\section*{Case Study: Evaluating Decision and Composite Risk in sNLI Systems on Choice-Overloaded Instances}\label{casestudy}
\begin{figure}[ht]
\centering
\includegraphics[width=0.7\textwidth]{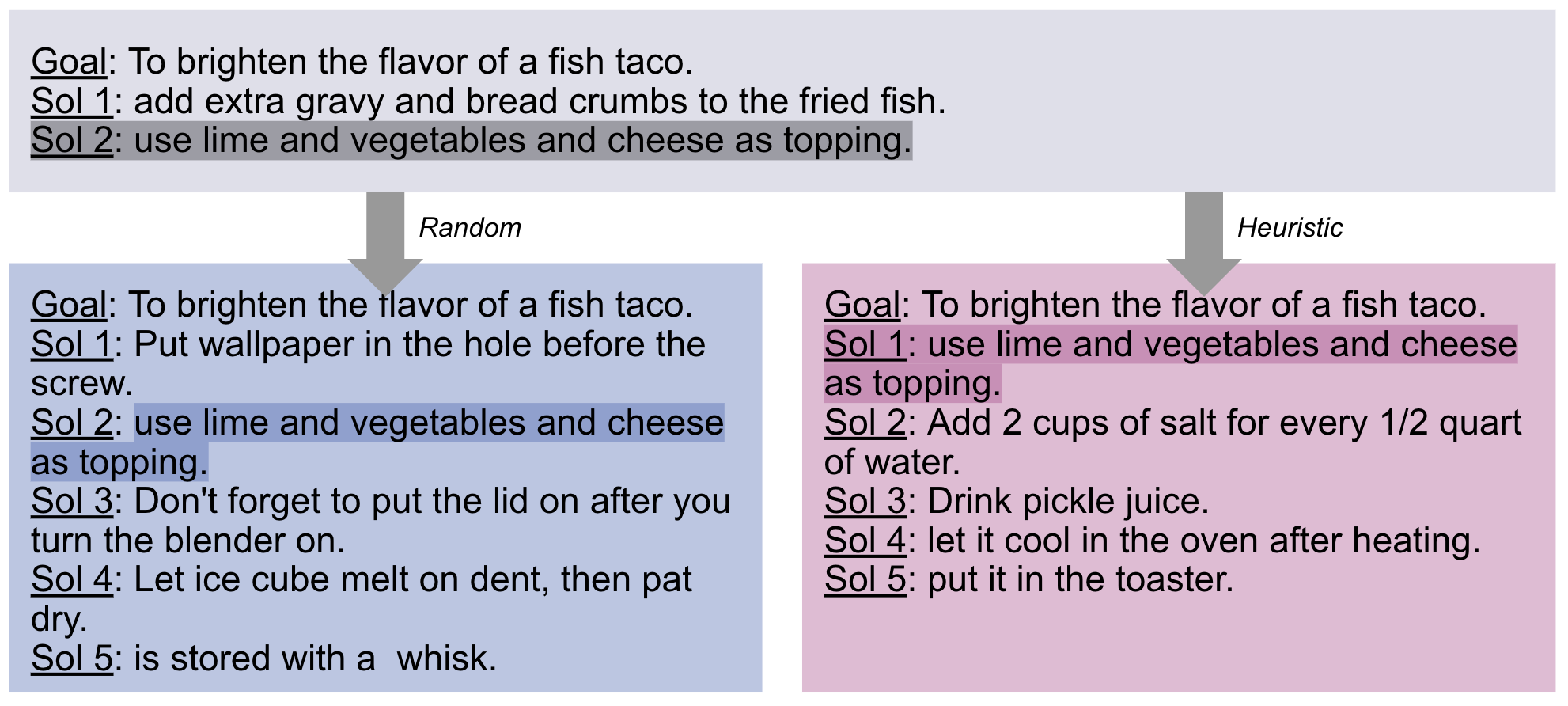}
\caption{Examples of choice overload in inference scenarios using random and heuristic sampling methods. The correct answers are highlighted.}
\label{fig:choiceParaExample}
\end{figure}

In previous sections, we presented RIFs as useful tools to simulate inference scenarios with injected risks, aiming to evaluate the ability of sNLI systems to detect and mitigate `artificial' risks. However, these simulated environments capture only a subset of the risks in real-world inference scenarios. Of particular interest is the `overload' effect—a cognitive bias where an excess of options leads to decision-making paralysis, impeding clear and rational judgment \cite{chernev2015choice,bensoussan2012analysis}. Our case study explores this issue by examining whether LMs perceive an `overloaded' context as intrinsically risky and if a decision rule can aid in identifying high-risk inferences in such scenarios. We put a spotlight on the DwD decision rule to evaluate its efficiency in detecting decision and composite risks in scenarios characterized by choice overload.

\begin{figure}[!ht]
\centering
\includegraphics[width=0.8\textwidth]{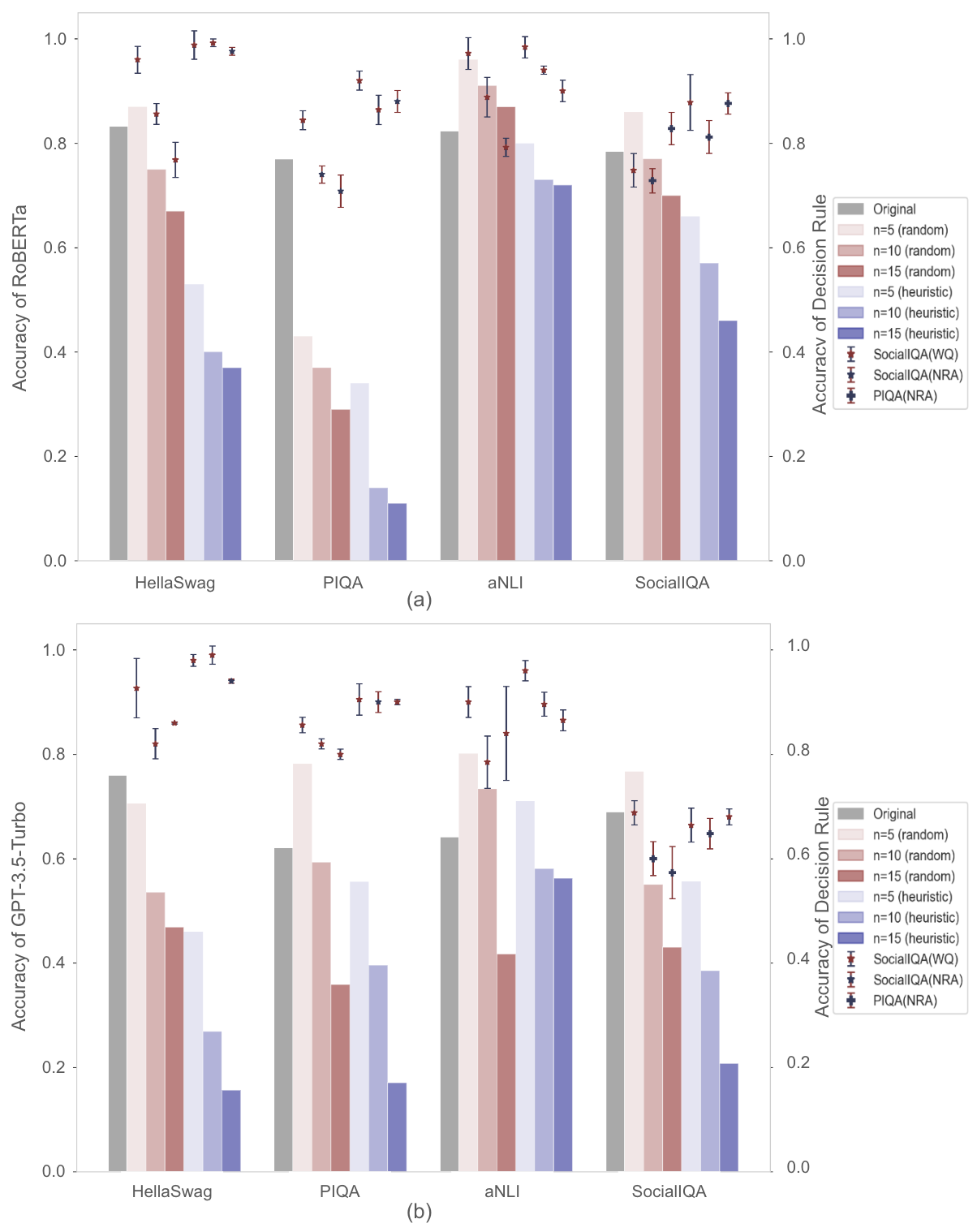}
\caption{Accuracy of RoBERTa (top) and GPT-3.5-Turbo (bottom) across four benchmarks under various choice paralysis settings. Error bars represent the performance range of the optimal DwD decision rule, which was trained on a training set from one of the benchmarks perturbed by one of the risk injection functions. Legend details include both choice paralysis settings and the corresponding top-performing DwD configurations.}
\label{fig:choicePara_acc_dr}
\end{figure}

\paragraph{Choice-overloaded Dataset Construction}

The development of choice-overloaded evaluation sets begins with a random selection of 50 instances from the original evaluation sets of the four previously mentioned benchmarks. These selected instances are then modified to expand their original set of candidate choices to specific predetermined numbers ($n = {5, 10, 15}$). To mitigate the influence of chance, for each specified number of candidate choices, we randomly selected 50 evaluation instances and repeated the experiment three times. Regardless of the benchmarks' original configuration of candidate choices -- for instance, HellaSwag offering four options and PIQA providing only two -- we retain only the original correct answer among the candidate choices. We then select $n-1$ candidate choices randomly from a collective pool of unrelated inference instances within the same benchmark. This pool is created by aggregating candidate choices from all instances, excluding the one undergoing modification, from which $n-1$ choices are drawn to complete the candidate set for each instance. These choices are subsequently shuffled to eliminate any bias in their ordering.

To explore how different incorrect choice sampling strategies impact the performance of sNLI systems on the constructed choice-overloaded datasets, we implement two distinct methodologies: the random method and the heuristic method. The random method randomly selects $n-1$ choices from the collective pool of candidate choices, ensuring a broad, albeit less targeted, selection. Conversely, the heuristic method employs a more nuanced approach by ranking all other inference instances according to the similarity of prompt embeddings across different inference instances to the target instance's prompt. It then identifies the $n-1$ instances most closely related semantically, and selects one incorrect answer from each to form a new set of candidate choices that are contextually similar. This heuristic approach aims to simulate a choice environment that not only overloads, but does so with options that are \textit{closely related }but still incorrect to the prompt, thus increasing the adversarial complexity and realism of the decision-making scenario. To ensure the validity of the candidate-choices expansion, we conducted a manual review to verify that each inference instance contained only one theoretically correct answer.  Figure \ref{fig:choiceParaExample} includes examples of choice-overloaded instances generated using both methods.

\paragraph{Decision Rule Method}

In the case study, we focus exclusively on the proposed DwD approach as the external decision rule. This set of DwD decision rules is consistent with those evaluated in prior decision and composite risk evaluations. The training of these DwD methods employs confidence scores from RoBERTa, yielded for a composite of instances from the initial training sets across the four benchmarks, alongside their corresponding ambiguous instances perturbed by a singular RIF (WQ or NRA). Hence, we explore eight distinct DwD decision rules, each serving as an advanced decision-making protocol for RoBERTa and GPT-3.5-Turbo within the sNLI systems.

\begin{figure}[!ht]
\centering
\includegraphics[height=\textwidth]{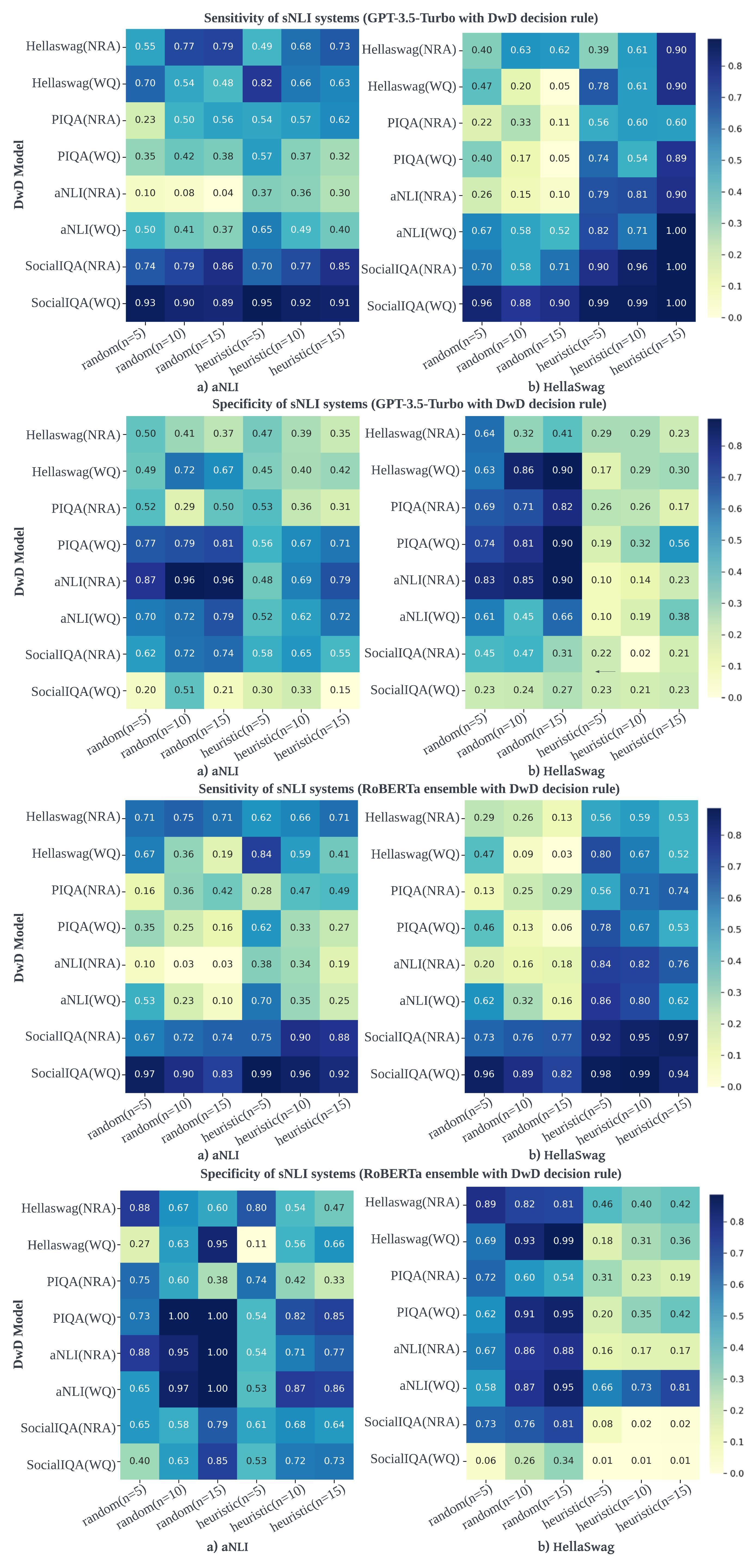}
\caption{Heatmaps of sensitivity and specificity in sNLI systems employing GPT-3.5-Turbo and RoBERTa, integrated with various DwD decision rules trained under distinct settings. The evaluations were performed across the aNLI (left) and HellaSwag (right) benchmarks with different choice paralysis settings.}
\label{fig:choicePara_sen_spe1}
\end{figure}

\begin{figure}[!ht]
\centering
\includegraphics[height=\textwidth]{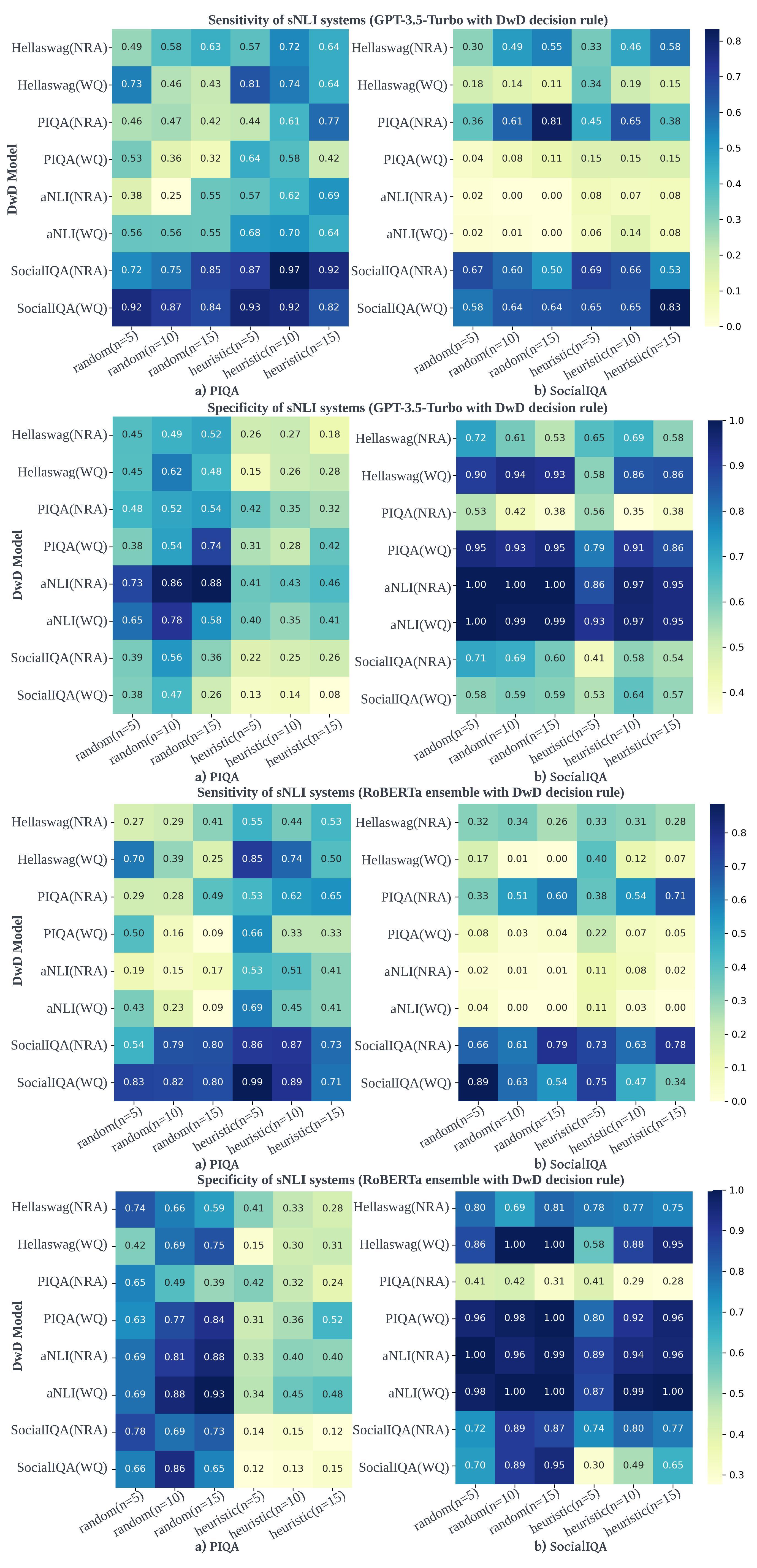}
\caption{Heatmaps of sensitivity and specificity in sNLI systems employing GPT-3.5-Turbo and RoBERTa, integrated with various DwD decision rules trained under distinct settings. The evaluations were performed across the PIQA (left) and SocialIQA (right) benchmarks with different choice paralysis settings.}
\label{fig:choicePara_sen_spe2}
\end{figure}
\paragraph{Results}

Figure \ref{fig:choicePara_acc_dr} presents the performance of both RoBERTa and ChatGPT on four benchmarks, comparing their accuracy across various choice-overloaded settings. Theoretically, LMs such as RoBERTa and CHatGPT should maintain consistent accuracy in selecting the correct answer, regardless of the number setting of options per instance. While a marginal decrease in accuracy might be expected in practice, it should not be significantly different from their performance on the original evaluation sets. In Figure \ref{fig:choicePara_acc_dr}, however, we find, interestingly enough, that both models exhibit improved performance on most benchmarks when the number of choices is increased to 5 with a random sampling method to extend the candidate choices. This suggests that the insertion of few random incorrect options may, counterintuitively, assist the models in more confidently identifying the correct answer across most benchmarks. As the number of options in the candidate choice sets increases, a clear decline in accuracy becomes apparent for both RoBERTa and ChatGPT, indicative of the models' struggles with an excess of options, which parallels the human experience of choice paralysis. This decline is particularly marked when the selection pool reaches 15, reflecting the substantial challenge these LMs face in identifying the correct answer from a broader selection of possibilities.

When the sampling method employed is the heuristic sampling, which is more adversarial since we are deliberately trying to confuse the model with an option that has a greater chance of being more related to the prompt and to the other answers, $n=5$ leads to a significant decline in performance for all benchmarks, with the exception of ChatGPT’s performance on aNLI. In fact, the results suggest that aNLI is the `easiest' benchmark for LMs when applying the choice overloaded expansion,  as it manages to remain within a 20\% margin from its baseline performance, even under the most aggressive setting ($n=15$). Even so, the results clearly illustrate that even on this benchmark, LMs are not immune from the choice paralysis problem.

Given that the correct answer is always included in the candidate choices for all choice-overloaded instances, an ideal decision rule would engage with each presented instance. The performance range of the top-performing DwD method, selected from the eight examined, is presented in Figure \ref{fig:choicePara_acc_dr}.  Our findings highlight that DwD instances, specifically those trained on the SocialIQA and PIQA benchmarks, demonstrate minimal decision risk across a variety of benchmarks under diverse choice expansion methods. These top DwD methods elect to respond to over 80\% of instances for RoBERTa and 75\% for GPT-3.5-Turbo, which is higher than LMs' accuracy on the original evaluation sets of the benchmarks. Yet, with the expansion of choice sets leading to a decrease in the accuracy of the LMs, DwD's frequency of responding to inference instances similarly decreases. To gain a deeper insight into the overall performance of the sNLI systems, we visualize the sensitivity and specificity of these systems, integrating the base LMs with different DwD rules, in heatmaps shown in Figures \ref{fig:choicePara_sen_spe1} and  \ref{fig:choicePara_sen_spe2} for the four benchmarks.

Figures \ref{fig:choicePara_sen_spe1} and \ref{fig:choicePara_sen_spe2} show that sNLI systems, when integrating GPT-3.5-Turbo with DwD methods trained on SocialIQA datasets perturbed with WQ or NRA RIFs, respond to an average of 80.8\% of instances where their selection mechanism is primed to deliver correct answers, regardless of the benchmark's choice overload settings. These sNLI systems exhibit their highest efficiency when employing DwD trained with WQ RIF-perturbed SocialIQA instances, achieving an average response rate of 91.8\% for correct predictions within OOD benchmarks including aNLI, HellaSwag, and PIQA. This engagement rate climbs notably when heuristic sampling, aimed at expanding candidate sets with more semantically related yet incorrect options, is implemented, suggesting the effectiveness of DwD in identifying instances where LMs can respond correctly. However, when confronting choice-overloaded sets derived from SocialIQA -- the ID evaluation set for the SocialIQA-trained DwD -- the sensitivity drops to approximately 0.6, suggesting the benchmark itself is a more challenging commonsense reasoning benchmark for LMs.

Within sNLI systems facing choice-overloaded inference, there's a notable trade-off between sensitivity and specificity. High sensitivity, indicative of the system's efficacy in recognizing true positives (instances can be correctly answered), typically has an inverse relationship with specificity, which measures the system's tendency to erroneously accept false positives.  However, sNLI systems employing DwD methods, particularly those trained on NRA RIF-perturbed SocialIQA instances, are able to withhold responses to 68\% of instances that would likely deliver erroneous answers within aNLI and SocialIQA evaluation sets expanded through random sampling. With a mean specificity of 0.4, sNLI systems containing the SocialIQA-trained DwD method abstain from answering 40\% of instances that were originally responded to incorrectly, while still engaging with over 90\% of those initially answered correctly. This careful selection boosts the accuracy, presenting the systems' robustness in managing risks, particularly in inference scenarios characterized by choice overload. Moreover, the exposure of DwD methods to RoBERTa's confidence score distributions grants sNLI systems that combine DwD with RoBERTa superior sensitivity and specificity over systems utilizing GPT-3.5-Turbo, elevating their accuracy rates further.

\section*{Discussion}  
\label{discussion}

To gain an intuitive sense of the inference scenarios in which composite risk arises, we list some representative questions in Tables \ref{table: correctDRnotanswer} and \ref{table: incorrectDRanswer}. Notable distinctions in composite risks are observed when comparing sNLI systems that employ ChatGPT against those based on RoBERTa. Specifically, ChatGPT-based sNLI systems tend to encounter composite risks in inference scenarios deemed `easier,' whereas RoBERTa-based systems face these risks in more complex inference contexts. These complex instances are often characterized by equivocality, challenging the system's ability to discern a singular correct answer, particularly in the presence of plausible distractors.

\begin{table}
\centering 
\begin{tabular}{lp{5.2cm}p{5.2cm}}  
\hline\hline
 & \textbf{GPT-3.5-Turbo} & \textbf{RoBERTa-Ensemble}  \\ 
\hline
\multirow{2}{*}{a\textbf{N}LI} & \begin{tabular}[c]{@{}l@{}}\uline{Obs 1}: Erin tried to learn how to draw. \\\uline{Obs 2}: So she joined a drawing class. \\\uline{Hyp 1}: Erin, practiced drawing at home\\with no luck. \\\uline{Hyp 2}: Erin, practiced drawing at home\\and became recognized for her talent.\end{tabular}  & \begin{tabular}[c]{@{}l@{}}\uline{Obs 1}: Sean was sitting at his desk. \\\uline{Obs 2}: After a minute, he was able to\\put the chair \\back together. \\\uline{Hyp 1}: He noticed the chair leg was\\falling off. \\\uline{Hyp 2}: He leaned too far back and his\\chair tipped over.\end{tabular}  \\ 
\cline{2-3}& \begin{tabular}[c]{@{}l@{}}\uline{Obs 1}: I used to procrastinate about\\studying. \\\uline{Obs 2}: Now, I never procrastinate\\studying. \\\uline{Hyp 1}: I failed a big test. \\\uline{Hyp 2}: After getting a good grade, I\\learned an easy lesson.\end{tabular}    & \begin{tabular}[c]{@{}l@{}}\uline{Obs 1}: Amy decided to move from\\Wisconsin to Florida. \\\uline{Obs 2}: However the experience was no\\fun without her \\friends. \\\uline{Hyp 1}: She would be with her friends\\out there. \\\uline{Hyp 2}: Amy wanted to live by the beach.\end{tabular}        \\ 
\hline
\multirow{2}{*}{\textbf{PIQA}} & \begin{tabular}[c]{@{}l@{}}\uline{Goal}: How do you properly prepare a\\steak. \\\uline{Sol 1}: Take the steak out of warm stor\\-age and let come to room temperature,\\generously add salt and pepper to both\\sides and let sit for 10 minutes. \\\uline{Sol 2}: Take the steak out of cold storage\\and let come to room temperature, \\generously add salt and pepper to both\\sides and let sit for 10 minutes.\end{tabular} & \begin{tabular}[c]{@{}l@{}}\uline{Goal}: To extend the shelf life of lettuce, \\\uline{Sol 1}: wrap it in paper towels before\\storing it in a bag. \\\uline{Sol 2}: wrap it in foil wrap instead of\\storing it in a bag.\end{tabular}                                                              \\ 
\cline{2-3}& \begin{tabular}[c]{@{}l@{}}\uline{Goal}: fire \\\uline{Sol 1}: can melt humans \\\uline{Sol 2}: can melt water\end{tabular}    & \begin{tabular}[c]{@{}l@{}}\uline{Goal}: Prevent earbuds from tangling. \\\uline{Sol 1}: Clip cord with hair clip. \\\uline{Sol 2}: Clip cord with hair tie.\end{tabular}  \\
\hhline{===}
\end{tabular}
\caption{Instances where LMs fail to provide correct answers, when DwD forces a response.}
\label{table: incorrectDRanswer}
\end{table}

\begin{table}
\centering
\begin{tabular}{lp{5.2cm}p{5.4cm}} 
\hline\hline
 & \textbf{GPT-3.5-Turbo}   & \textbf{RoBERTa-Ensemble} \\ 
\hline
\multirow{2}{*}{\textbf{aNLI}} & \begin{tabular}[c]{@{}l@{}}\uline{Obs 1}: Trevor went to the lake one day\\to fish.\\\uline{Obs 2}: Trevor was forced to go home\\after he lost \\his fishing pole. \\\uline{Hyp 1}: The water was perfect for all\\levels of fishing. \\\uline{Hyp 2}: The water was spitting up poles.\end{tabular}                                        & \begin{tabular}[c]{@{}l@{}}\uline{Obs 1}: Amy and her friends were out\\at 3 AM. \\\uline{Obs 2}: They stayed there breathing hard\\and praying they hadn't been seen. \\\uline{Hyp 1}: They started getting followed by\\a policeman, ran, and hid behind a build\\-ing.\\\uline{Hyp 2}: They decided to break into the\\football field. When suddenly they saw\\a flashlight comming towards them. \\They all started running for the bleachers.\end{tabular}  \\ 
\cline{2-3} & \begin{tabular}[c]{@{}l@{}}\uline{Obs 1}: My four-year-old nephew loves\\to wake us up.\\\uline{Obs 2}: As I screamed, he yelled cold\\hands.\\\uline{Hyp 1}: Today I was ready for him.\\When he came into our room a jumped\\out and tickled him.\\\uline{Hyp 2}: He would jump on our ear to\\get our attention.\end{tabular}             & \begin{tabular}[c]{@{}l@{}}\uline{Obs 1}: My four-year-old nephew loves to\\wake us up.\\\uline{Obs 2}: As I screamed, he yelled cold\\hands.\\\uline{Hyp 1}: Today I was ready for him. When\\he came into our room a jumped out and\\tickled him.\\\uline{Hyp 2}: He would jump on our ear to get\\our attention.\end{tabular}   \\ 
\hline
\multirow{2}{*}{\textbf{PIQA}} & \begin{tabular}[c]{@{}l@{}}\uline{Goal}: How to sneak past guards in\\Metal Gear NES?\\\uline{Sol 1}: Wait until they say "I feel asleep!"\\and walk past, or wait until they face\\another direction. \\\uline{Sol 2}: Wait until they say "I'm feeling\\sleepy!" and walk past, or wait until\\they face another direction.\end{tabular} & \begin{tabular}[c]{@{}l@{}}\uline{Goal}: To cook perfectly golden pancakes,\\\uline{Sol 1}: keep the temperature high and cook\\quickly.\\\uline{Sol 2}: keep the temperature low for a long\\-er time.\end{tabular}      \\ 
\cline{2-3}& \begin{tabular}[c]{@{}l@{}}\uline{Goal}: How can I make a funnel?\\\uline{Sol 1}: Cut an empty plastic soda bottle\\in half and use the top.\\\uline{Sol 2}: Cut an empty plastic soda bottle\\in half and use the bottom.\end{tabular}                                                                                                 & \begin{tabular}[c]{@{}l@{}}\uline{Goal}: How can I make a funnel?\\\uline{Sol 1}: Cut an empty plastic soda bottle in\\half and use the top.\\\uline{Sol 2}: Cut an empty plastic soda bottle\\in half and use the bottom.\end{tabular}   \\
\hhline{===}
\end{tabular}
\caption{Instances where LMs can provide correct answers when DwD abstains from answering.}
\label{table: correctDRnotanswer}
\end{table}

Table \ref{table: correctDRnotanswer} presents instances where, despite the selection rule's capacity to offer correct responses, the decision rule withhold the answer. There are more instances where distinct sNLI systems, employing various LMs, concurrently encounter composite risks. Additionally, while benchmarks such as PIQA are designated as measures of commonsense reasoning, some of their content, such as questions about the game ``Metal Gear NES,'' may not align well with the typical understanding of commonsense. This variation indicates the importance of a more thorough manual examination of the instances included in these benchmarks.

Furthermore, the recurrence of equivocality in certain instances indicates room for improvement in decision rule methods. The DwD approach, which introduces artificially ambiguous examples lacking a clear-cut correct answer, may lead the system to prioritize the detection of ambiguity over the actual answerability, leading to suboptimal performance. As this research is centered around a risk-centric evaluation framework, our future work will focus on developing more advanced decision rule methods. These methods will aim for a more effective calibration based on the confidence yielded by LMs and to align more closely with the LM's understanding of an instance and the rationale behind the decision rule's choice to engage or not.

\section*{Conclusion} \label{conclusion}
This paper proposed and applied a risk-centric evaluation framework that defined two types of risk: decision and composite risk. 
Using four NLI benchmarks, we conducted an experimental study to demonstrate the practical utility of the proposed framework. A key finding of the study is that less well-performing confidence calibration can lead to problems of both under-confidence and over-confidence, despite considerably greater attention given to the latter in the literature. Learning-based decision rules, such as DwD, can help such models minimize the risks, even in challenging inference situations such as choice overload, while maintaining good overall inference performance.

\bibliography{sample}

\section*{Acknowledgements}

This work was funded under the DARPA Machine Common Sense (MCS) program under award number N660011924033.

\section*{Author contributions statement}

K.S. conceived and conducted all experiments, analysed the results, and wrote the manuscript. M.K. supervised the project, and edited the manuscript. All authors reviewed the manuscript.

\section*{Data availability statement}

All data generated or analysed during this study are included in this published article [and its supplementary information files].

\section*{Additional information}


\textbf{Accession codes:} Not Applicable; \textbf{Competing interests:} The authors have no competing interests to declare.

\end{document}